%% file: main.tex
\documentclass[10pt,twocolumn,letterpaper]{article}

\input{style}

\begin{document}


\title{Diagnosing Rarity in Human-object Interaction Detection}

\author{Mert Kilickaya\\
QUvA Deep Vision Lab\\
Amsterdam, Netherlands\\
{\tt\small m.kilickaya@uva.nl}
\and
Arnold Smeulders\\
QUvA Deep Vision Lab\\
Amsterdam, Netherlands\\
{\tt\small a.w.m.smeulders@uva.nl}
}

\maketitle

\begin{abstract}

Human-object interaction (HOI) detection is a core task in computer vision. The goal is to localize all human-object pairs and recognize their interactions. An interaction defined by a \texttt{<verb, noun>} tuple leads to a long-tailed visual recognition challenge since many combinations are rarely represented. The performance of the proposed models is limited especially for the tail categories, but little has been done to understand the reason. To that end, in this paper, we propose to diagnose rarity in HOI detection. We propose a three-step strategy, namely Detection, Identification and Recognition where we carefully analyse the limiting factors by studying state-of-the-art models. Our findings indicate that detection and identification steps are altered by the interaction signals like occlusion and relative location, as a result limiting the recognition accuracy.  

\end{abstract}

\input{0-introduction}

\input{2-method}
\input{3-experiment}
\input{4-conclusion}

{\small
\bibliographystyle{ieee_fullname}
\bibliography{egbib}
}

\end{document}

%% file: style.tex
\usepackage{cvpr}
\usepackage{times}
\usepackage{epsfig}
\usepackage{graphicx}
\usepackage{amsmath}
\usepackage{amssymb}
\usepackage{caption}
\usepackage[nocompress]{cite}
\usepackage{slashbox}
\usepackage[dvipsnames,table]{xcolor} 
\usepackage[toc,page]{appendix}
\usepackage{enumerate}
\usepackage{tabularx,ragged2e}
\usepackage{spreadtab}
\usepackage{adjustbox}
\usepackage{booktabs}
\usepackage{xcolor}
\usepackage{multirow}
\usepackage{rotating}
\usepackage{color, colortbl}
\usepackage{subfigure}

\usepackage[inline]{enumitem}

\usepackage[pagebackref=true,breaklinks=true,letterpaper=true,colorlinks,bookmarks=false]{hyperref}

\graphicspath{{./figures/}}

\DeclareGraphicsExtensions{.jpeg,.png,.eps,.pdf}

\definecolor{Gray}{gray}{0.9}

\definecolor{Gray}{gray}{0.85}
\definecolor{LightCyan}{rgb}{0.88,1,1}

\newcolumntype{a}{>{\columncolor{Gray}}c}
\newcolumntype{b}{>{\columncolor{white}}c}

\newcommand{\partitle}[1]{\bigbreak\noindent\textbf{#1}}

\makeatletter
\newcommand*{\rom}[1]{\expandafter\@slowromancap\romannumeral #1@}
\makeatother

\cvprfinalcopy 

\def\cvprPaperID{3} 


%% file: 0-introduction.tex
\section{Introduction}

\begin{figure}[t]
\begin{center}

\includegraphics[width=0.9\linewidth]{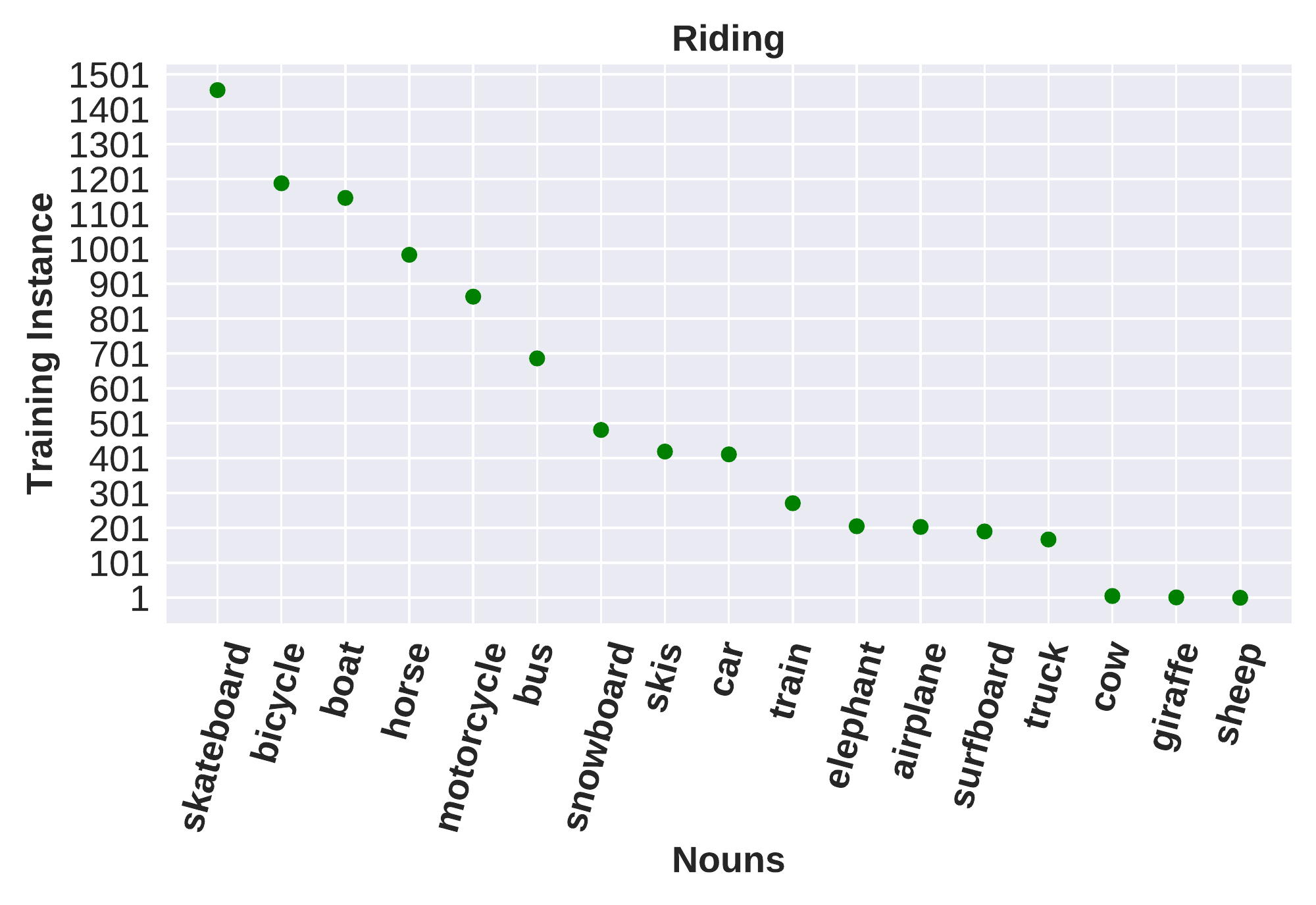}
\end{center}
  \caption{Human-object interactions exhibit a long-tail. For example, riding skateboard has more than $1k$ training instances whereas cow, giraffe or sheep only has $1$ example.}
  \label{fig:teaser}
\end{figure}

\noindent The goal of HOI detection is to detect all possible human-object pairs and recognize their interactions from an image. It is a core task in computer vision with many applications in robotics~\cite{katz2008manipulating}. A HOI is defined by a triplet of \texttt{<human, interaction, object>}, where the human and the object is a bounding box and the interaction is a \texttt{<verb, noun>} pair, such as \texttt{<ride, bicycle>}. The task received an increasing amount of attention in recent years~\cite{gkioxari2018detecting,qi2018learning,gao2018ican,xu2019learning,peyre2018detecting,gupta2018no,bansal2019detecting,wang2019deep,chafik2020classifying} thanks to the benchmark dataset HICO-Det~\cite{chao2018learning}. The distribution of the training samples for interactions follows a long-tailed distribution where many interactions have few examples, see Figure~\ref{fig:teaser}. Despite the progress in the performance of many-shot interactions, recognizing rare interactions remains a challenge.  

\begin{figure*}[h]
\includegraphics[width=1.0\linewidth]{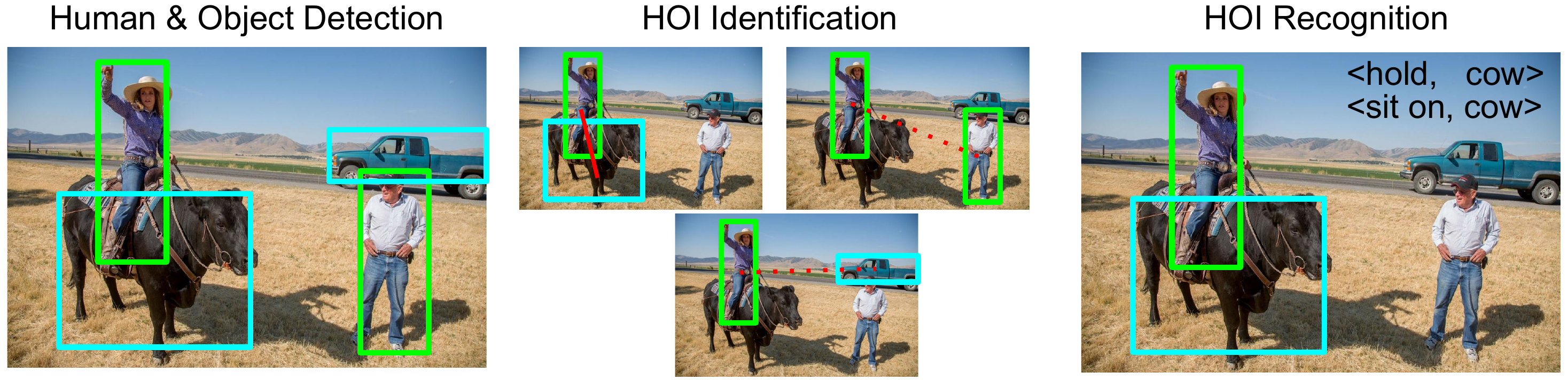}
  \caption{Steps of our diagnostic study. Detection (left), Identification (middle) and Recognition (right).}
  \label{fig:steps}
\end{figure*}

HOI Detection is accomplished in three steps, see Figure~\ref{fig:steps}. \textbf{1. Human-object detection}: HOI detector initially localizes all possible human-objects from the image. This step is challenged by the fact that interactions transform the human-object appearance, such as occlusions due to grasping, making it hard to localize all human-objects. \textbf{2. HOI Identification}: Given the exhaustive pairing of all detected humans and objects, the HOI detector needs to identify the real interacting pairs. In the case of Figure~\ref{fig:steps}, only the rider, and the cow are in an interaction. This step is challenging since the cues of an interacting pair is in their subtlety, such as the relative locations of human-objects, human gaze or object parts. \textbf{3. HOI Recognition}: In this step, HOI detector needs to classify the interaction type of the human-object pair(s), such as \texttt{<hold, cow>} and \texttt{<sit on, cow>} in Figure~\ref{fig:steps}. It is hard to distinguish among many interaction types since only a few examples are available for many interactions. 


Existing models highlight the difficulty in recognizing rare HOIs by reporting the performance on both the rare and nonrare (\ie many shots) splits on the benchmark dataset~\cite{chao2018learning}. However, it is not known what makes rare interactions particularly challenging aside from the low number of examples. Are the human-objects of rare interactions harder to detect? Or rare interacting pairs are harder to identify? The goal of our paper is to answer these questions. Instead of engineering a new model, we try to understand the detectability and identifiability of rare interactions with the help of the state-of-the-art HOI detectors~\cite{gao2018ican,chao2018learning,li2019transferable}. 

Our findings are: (i) Localizing human-objects of rare interactions is not challenging, however, detection is altered by the small and occluded human-objects, (ii) Identifying the rare HOIs is challenging and is altered by the background clutter and human-object distance, and (iii) Recognizing rare interactions is influenced by the detection and identification errors, leaving a big room for improvement.

%% file: 2-method.tex
\section{Empirical Material}


\subsection{Benchmark Dataset}

For our analysis, we resort to HICO-Det dataset~\cite{chao2018learning}. HICO-Det is the biggest HOI detection benchmark, with a diverse set of categories. The dataset comes with (i) Human-object bounding boxes for detection, (ii) Human-object interaction pair annotations for identification, and (iii) Human-object interaction types for recognition. A typical human-object concurrently performs multiple interactions such as holding, sitting on and riding a bicycle which is exhaustively annotated. The dataset has in total $47k$ number of images, with more than $150k$ human-object pair annotations. There exists $600$ distinct interaction types of which $168$ are rare, for $80$ unique nouns and $117$ unique verbs. 

A unique property of the dataset is, for each noun in the dataset, there exists a no-interaction category, where at least a human and the target object is in the image, though not performing an interaction(\eg the man and the car in Figure~\ref{fig:steps}). This enforces the models to focus on the interaction characteristics as opposed to leveraging human-object co-occurrence. 

\subsection{Benchmark HOI Detectors}

For our diagnostic purposes, we resort to three state-of-the-art HOI detectors, namely HO-RCNN~\cite{chao2018learning}, iCAN~\cite{gao2018ican}, and TIN~\cite{li2019transferable} for the following reasons: (i) Their high performance, (ii) Standard use of the same object detector, the same backbone and the same number of layers and (iii) Publicly available code. Here, we present an overview.


\partitle{HO-RCNN~\cite{chao2018learning}.} HO-RCNN is a three-stream Convolutional Neural Network. Each stream considers the appearance of either the human, the object or the human-object (pairwise). Human and object streams consider the global appearance of human and object regions obtained via Region-of-Interest pooling~\cite{fastrcnn}, whereas pairwise stream considers the spatial layout of human-object locations. Spatial locations are critical especially to identify a possible interaction between a human and an object. The detector is built upon the Faster-RCNN backbone~\cite{faster-rcnn}. For human-object detection, the model makes use of this backbone pre-trained on MS-coco~\cite{mscoco}. HOI recognition is achieved by combining the individual predictions of the three streams. 

\partitle{iCAN~\cite{gao2018ican}.} iCAN follows the same network structure as HO-RCNN, and couples HO-RCNN network with a self-attention mechanism~\cite{nonlocal} called instance-centric attention layer for the human and the object streams. This highlights the fine-grained details within human-object regions that are essential to HOIs.  

\partitle{TIN~\cite{li2019transferable}.} TIN augments iCAN with an interactivity classifier, that predicts whether if a given pair of human-object are in interaction or not. The authors prune (suppress) those pairs that are predicted to be non-interacting by the interactivity classifier prior to recognition. Since it is critical to identify which pairs are interacting before the recognition, the authors obtain a considerable improvement in both rare and nonrare interactions over iCAN.  

\partitle{Implementation details.} All the models are trained for $1.8$ million iterations with image-based training~\cite{faster-rcnn}. The learning rate is set to $0.001$ decayed after $900k$ iterations. Weight decay ($0.0001$), dropout (with keep probability $p=0.4$) and batch normalization~\cite{batchnorm} is used for regularization. We include all human detections with a confidence higher than $0.8$ and all object detections with a confidence higher than $0.3$. Using this standard gave a boost to HO-RCNN as it yields better results than iCAN in our experiments. 

%% file: 3-experiment.tex
\section{Diagnostic Analysis}

We diagnose the HOI detection in three steps, namely Human-object detection, HOI identification, and HOI recognition. We use the standard rare/nonrare split proposed in~\cite{chao2018learning}. 

\subsection{Detection of Human and Object}

To diagnose human and object detection, we first measure the recall of the off-the-shelf detector Faster-RCNN~\cite{faster-rcnn} commonly used by all three models. We measure the recall using the traditional PASCAL VOC criteria~\cite{pascalvoc} that Intersection-over-Union IoU $\textgreater 0.50$  over the full, the rare and the nonrare splits. Results can be seen from Table~\ref{tab:exp11}.

\begin{table}[h]

\begin{center}
\setlength\tabcolsep{8.5pt}
\begin{tabular}{lccccc}

\toprule
   & Full & NonRare & Rare \\
\midrule 

Human  & $86$ & $86$ & $92$ \\ 
Object & $63$ & $64$ & $90$ \\ 
\midrule 

\end{tabular}
   \caption{Recall of human and object detections.}
\label{tab:exp11}
\end{center}
\end{table}


Results show that there is no gap in terms of human and object detection performance between rare and nonrare categories. This indicates that recognizing rare interactions is not altered by the detection step. 

Then, we study the sensitivity of the detector to the area and the occlusion. We measure the area of a bounding box by the width $\times$ height of the box divided by the number of pixels in the image. We measure the occlusion of box $b_j$ on $b_i$ as $\dfrac{b_{i} \cap b_{i}}{area(b_{j})}$~\cite{vezhnevets2015object}. An area or occlusion is small if $\textless 0.20$. Results can be seen from Table~\ref{tab:exp12}.

\begin{table}[h]

\begin{center}
\setlength\tabcolsep{6.5pt}
\begin{tabular}{lccccc}

\toprule
   & Full     & \multicolumn{2}{c}{Area} & \multicolumn{2}{c}{Occlusion} \\ 
   &  & Small & Bigger & Small & Bigger \\
 \midrule
Human  & $86$ & $44$ & $94$ & $95$ & $66$ \\ 
Object & $63$ & $26$ & $79$ & $84$ & $43$ \\ 
   
\midrule 

\end{tabular}
   \caption{Sensitivity of human-object detection.}
\label{tab:exp12}
\end{center}

\end{table}

\begin{figure}[h]
\begin{center}
\includegraphics[width=1.00\linewidth]{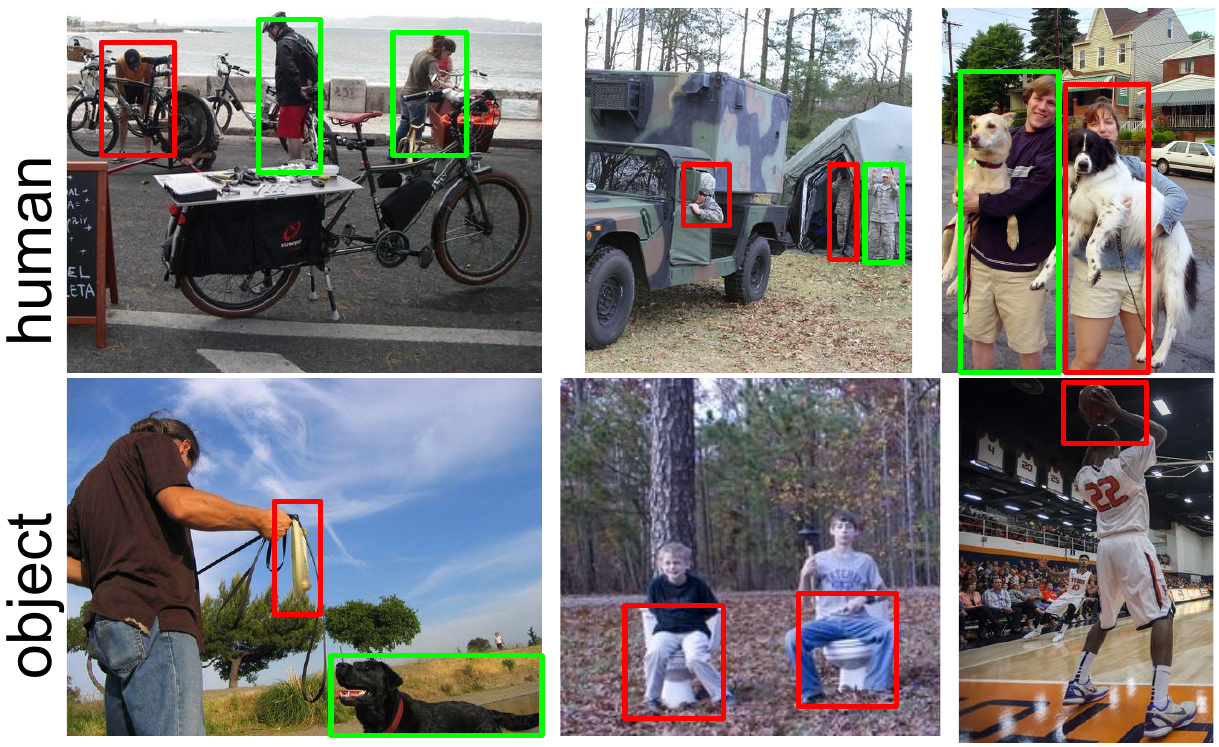}
\end{center}
  \caption{Example human-object detector \textcolor{green}{success} and \textcolor{red}{failure} cases.}
  \label{fig:detector}
\end{figure}

Results indicate that the off-the-shelf detector is sensitive to both the small area and the large occlusion of humans and objects. This can limit the performance in subsequent steps since many human-object interactors occupy a small region in the image and is occluded by each other. We give detection examples in Figure~\ref{fig:detector} for the human and the objects (green for detected, red for missed boxes). Observe how, within the same image, the detector fails to localize the humans or the objects that have bigger occlusions or occupy a small region.   

It is concluded that localization of humans and objects is not altered for rare interactions, however, is affected by the area and the occlusion on human and object regions.

\subsection{Identification of Human-Object Interaction}

To diagnose HOI Identification, we measure the binary accuracy of identification (interacting vs. not-interacting) performance. To obtain interaction vs. not-interaction scores from each model, we obtain maximum response over all interaction ($520$ classes) and not-interaction ($80$ classes) categories respectively. Results are in Table~\ref{tab:exp21}. 

\begin{table}[h]

\begin{center}
\setlength\tabcolsep{8.5pt}
\begin{tabular}{lccccc}

\toprule
   & Full & NonRare & Rare \\
\midrule 

HO-RCNN~\cite{chao2018learning} & $79$ & $79$ & $71$ \\

iCAN~\cite{gao2018ican} & $74$ & $75$ & $67$ \\

TIN~\cite{li2019transferable} & $82$ & $82$ & $77$  \\
\midrule 

\end{tabular}
   \caption{Identification accuracy of human-object interaction for rare and non-rare interaction categories.}
\label{tab:exp21}
\end{center}
\end{table}

Results reveal that for all three models there is a consistent gap in the identification performance between rare and nonrare interactions. This indicates that the models demand more training examples to learn interactivity. 

We then study the sensitivity of human-object distance and human-object clutter in identification accuracy. We compute human-object distance as the number of pixels between the centers of humans and objects divided by the number of pixels in the image. We compute human-object clutter as the number of human-objects in the background (\ie not involved in any interaction). The distance is deemed to be small if $\textless 0.20$ and the clutter deemed to be small if less than $5$ other human-objects exist in the background. Results can be seen from Table~\ref{tab:exp22}.

\begin{table}[h]

\begin{center}
\setlength\tabcolsep{4.5pt}
\begin{tabular}{lccccc}

\toprule
   & Full     & \multicolumn{2}{c}{Distance} & \multicolumn{2}{c}{Clutter} \\ 
   &  & Small & Bigger & Small & Bigger \\
 \midrule
HO-RCNN~\cite{chao2018learning}  & $79$ & $94$ & $77$ & $91$ & $69$ \\ 
iCAN~\cite{gao2018ican}    & $74$ & $91$ & $72$ & $88$ & $65$ \\ 
TIN~\cite{li2019transferable}     & $82$ & $96$ & $80$ & $93$ & $74$ \\ 
      
\midrule 

\end{tabular}
   \caption{Sensitivity of identification.}
\label{tab:exp22}
\end{center}

\end{table}


Results indicate that HOI identification is challenged by human-object distance. The models tend to assign distant human-objects to not-interaction category, leading to errors in identification. This shows that the models mostly leverage the spatial layout of human and object locations to identify the interaction. 

It is also clear that background objects confuse the HOI identification step. As the number of possible human-object pairs increases with the background human-objects, the classifier finds it hard to distinguish interacting pairs from non-interacting ones. 

\begin{figure}[h]
\begin{center}
\includegraphics[width=1.00\linewidth]{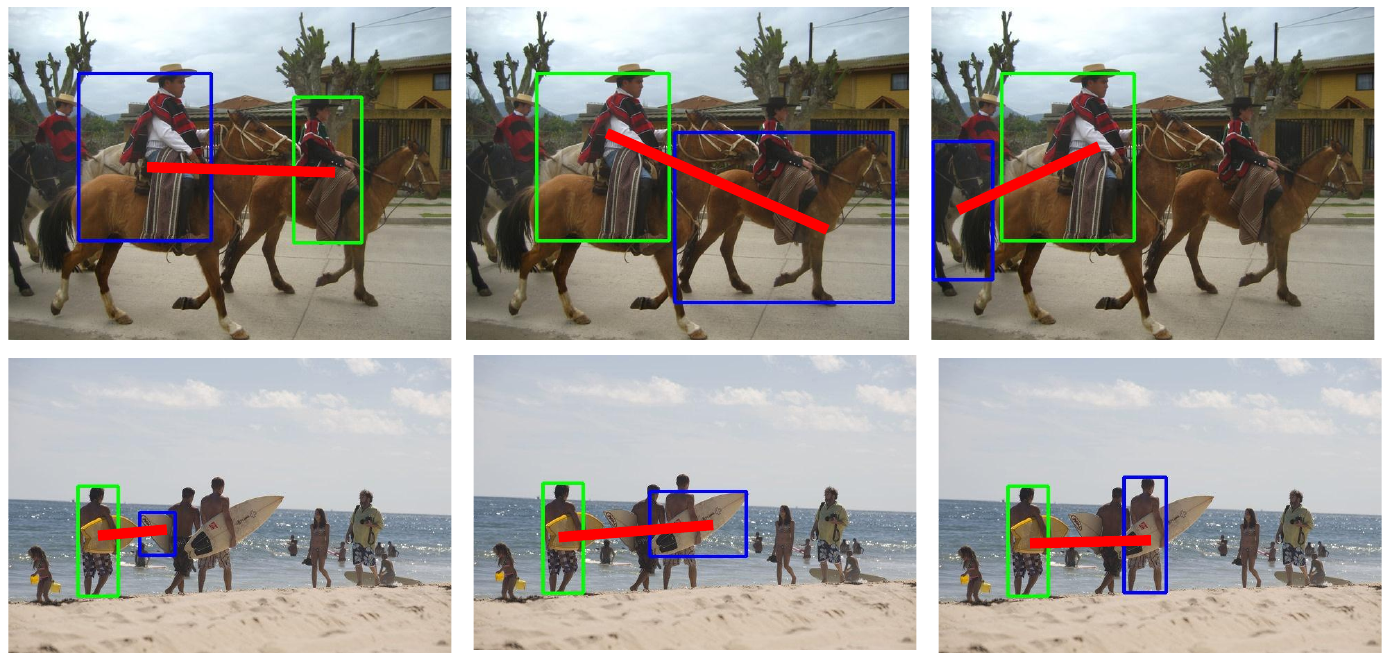}
\end{center}
  \caption{Failure examples for HOI Identification of TIN~\cite{li2019transferable}. The rider (above) and the surfer (below) is paired with three other surrounding objects even though there is no interaction in between, lowering the identification performance.}
  \label{fig:identification}
\end{figure}

The effect of the human-object distance and the human-object clutter on identification is visualized in Figure~\ref{fig:identification} for TIN~\cite{li2019transferable}. In both images there are multiple humans and multiple objects, making it a real test for HOI identification. The rider (above) and the surfer (below) are paired with three other surrounding objects with even though there is no interaction in between, lowering the identification performance. 

It is concluded that HOI identification is inferior for the rare interaction categories, and HOI identification is altered by the existence of the human-object distance and human-object clutter.

\subsection{Recognition of Human-Object Interaction}







To diagnose HOI recognition, in line with detection and identification steps, we measure the overall performance at the human-object instance-level. Specifically, we compute the mean Average Precision of interaction classification performance, which is then averaged over all pairs (not over classes) over the dataset. The results are presented in Table~\ref{tab:exp31}. 

\begin{table}[h]

\begin{center}
\setlength\tabcolsep{8.5pt}
\begin{tabular}{lccccc}

\toprule
   & Full & NonRare & Rare \\
\midrule 

HO-RCNN~\cite{chao2018learning} & $9$ & $10$ & $1$ \\

iCAN~\cite{gao2018ican} & $8$ & $9$ & $1$ \\

TIN~\cite{li2019transferable} & $10$ & $10$ & $2$  \\
\midrule 

\end{tabular}
   \caption{Recognition mAP for rare and non-rare interaction categories.}
\label{tab:exp31}
\end{center}
\end{table}

Results are in line with the class-based mAP reported in all three models ~\cite{chao2018learning,gao2018ican,li2019transferable}, as there is a big gap in recognizing rare and nonrare interaction instances. We then measure the sensitivity of recognition to the detection and identification errors. A detection is correct if IoU $\textgreater 0.50$, and incorrect otherwise. An identification is correct if the accuracy is $1$. Results can be seen from Table~\ref{tab:exp32}.

\begin{table}[h]

\begin{center}
\setlength\tabcolsep{3.5pt}
\begin{tabular}{lccccc}

\toprule
   & Full     & \multicolumn{2}{c}{Detection} & \multicolumn{2}{c}{Identification} \\ 
   &  & Incorrect & Correct & Incorrect & Correct \\
 \midrule
HO-RCNN~\cite{chao2018learning}  & $9$ & $3$ & $14$ & $3$ & $16$ \\ 
iCAN~\cite{gao2018ican}    & $8$ & $3$ & $13$ & $4$ & $13$ \\ 
TIN~\cite{li2019transferable}     & $10$ & $4$ & $14$ & $4$ & $23$ \\ 
\midrule 

\end{tabular}
   \caption{Sensitivity of recognition.}
\label{tab:exp32}
\end{center}

\end{table}

Results indicate that both the detection and identification errors alter the recognition performance as expected. However, a correct detection or identification does not guarantee perfect recognition accuracy, indicating that the human-object representation is not discriminative. 

%% file: 4-conclusion.tex
\section{Conclusion}


This paper focused on rarity in HOI detection in three steps. We revealed that human-object detection is altered by occlusions and area, that are abundant in HOIs. This calls for interaction-specific human-object detectors since for a conventional detector occlusion is a nuisance rather than a signal. We also showed that identifying rare HOIs is difficult which calls for finer-grained reasoning beyond spatial layout. Lastly, we show that recognition is limited by detection and identification errors, which leaves a big room for improvement for these specific steps. 